# POLYHIERARCHICAL CLASSIFICATIONS INDUCED BY CRITERIA POLYHIERARCHIES, AND TAXONOMY ALGEBRA

Pavel Babikov, Oleg Gontcharov, and Maria Babikova
QNT Software Development Inc.
528 Victoria Ave, Windsor, ON N9A 4M8, Canada
Email: {babikov,gontcharov,mbabikova}@quasinewtonian.com

## Abstract

A new approach to the construction of general persistent polyhierarchical classifications is proposed. It is based on implicit description of category polyhierarchy by a generating polyhierarchy of classification criteria. Similarly to existing approaches, the classification categories are defined by logical functions encoded by attributive expressions. However, the generating hierarchy explicitly predefines domains of criteria applicability, and the semantics of relations between categories is invariant to changes in the universe composition, extending variety of criteria, and increasing their cardinalities. The generating polyhierarchy is an independent, compact, portable, and re-usable information structure serving as a template classification. It can be associated with one or more particular sets of objects, included in more general classifications as a standard component, or used as a prototype for more comprehensive classifications.

The approach dramatically simplifies development and unplanned modifications of persistent hierarchical classifications compared with tree, DAG, and faceted schemes. It can be efficiently implemented in common DBMS, while considerably reducing amount of computer resources required for storage, maintenance, and use of complex polyhierarchies.

## 1. Introduction

The majority of common classification schemes are based on hierarchies of categories represented in terms of classic trees (see [1]). As it is well known, the principal disadvantages of trees are 1) the impossibility to use several classification criteria concurrently in the process of concretization, thereby leading to a necessity of pre-ordering criteria by significance or rank (the «predefined path problem»), and 2) catastrophic multiplication of identical criteria in parallel sub-trees (the «sub-trees multiplication problem»). So the recent years show a boost in the development of alternative classification schemes, including compositions of trees and set-theory methods of identifying categories. Some of the most successful new schemes are faceted classifications [2] and the formalisms used in rough sets theory [3] and granular computing [4]. However, these schemes, while successful in specific fields of application, do not resolve all major problems related to the construction of general polyhierarchical taxonomies.

The faceted classification schemes are based on concurrent use of a number of separate hierarchies (facets), and include a formalism of relations between them. They provide simultaneous access to the classification criteria pertaining to different facets, and implicitly define categories in terms of compositions of independent classifications without explicitly enumerating them. However, some serious problems still remain. In practical cases it can be difficult or even impossible to split the target classification into a set of independent hierarchies rooting from the whole universe. This results in complication of facet internal structures, thus returning back to the predefined path and sub-trees multiplication problems. On the other hand, separating re-usable sub-trees requires introduction of respective «conditional» facets rooting from particular subsets. That leads to the problem of automatic recognition of facets applicability and consistency, which requires introduction of supplementary descriptive structures, such as purposes and roles of facets, meta-facets and the like. When developing a full-scale practical classification this system of auxiliary entities may become too sophisticated for programming and maintenance, not saying about its mathematical inelegance and excessive involvement of heuristic considerations.

In addition, the faceted schemes do not directly support systemizing of categories by their generality which is necessary for the classification to be a persistent polyhierarchy and for an efficient work with abstract categories. Introduction of locally applicable facets (sub-facets) while keeping hierarchical structure of the entire classification generally causes further encumbering of descriptive structures. To avoid this problem many present-day faceted schemes even do not care of automatic recognition of domains of criteria applicability (see, e.g. [5]), thus creating a lot of room for errors when developing and using classification. Therefore, the faceted schemes can be an excellent tool for knowledge representation in terms of relations between relatively simple independent hierarchies, but not so for the cases of very general persistent polyhierarchical classifications..

Within the formalism used in rough sets and granular computing, the categories of classification are defined by logical expressions based on a system of attributes describing object properties. Such a description allows an efficient execution of set theory operations on categories while maintaining the hierarchical structure of the classification. However, both the resulting definitions of classes (categories) in terms of logical expressions and relations between them depend on a composition of the information table, i.e. they are not invariant with respect to modifications of the classification universe. Due to the classic tree structuring the decision rule hierarchies are subject to both the predefined path and sub-tree multiplication problems. Also, that formalism is not adapted for cases when some attributes have meaning only for particular subsets of the available objects. So, the applicability of the rough sets and granular computing formalism is limited by the automatic construction of empirical classifications of given sets of objects described in terms of globally meaningful properties.

In this paper a new descriptive formalism is proposed that combines the advantages of other classification schemes. It has been designed as a general-purpose tool for building complex multi-criteria classifications, while satisfying the following requirements:

1. Existence of a global polyhierarchical structure directly and inherently supporting recognition of domains of criteria applicability and providing simultaneous (random) access to all the applicable criteria.
2. Persistence of the polyhierarchy and, in particular, invariance of its previously developed parts with respect to extension of the universe, addition of new options to existing criteria, and introduction of new classification criteria.
3. Compactness of descriptive data structures allowing to avoid cumulative multiplication of explicitly enumerated classification categories and explicitly described relations between them.
4. Support of set-theoretic operations, including intersections, unifications and complements (differences) of categories.
5. Efficient realization of the test for distant inheritance relationships between categories.
6. Conceptual simplicity of the design process, as well as further unplanned extensions and refinements.

Our approach is based on the introduction of a partially ordered system of classification criteria in such a way that domain of definition, i.e. area of applicability of each criterion is explicitly defined by composition of classifications by some more general criteria (if any). Thus, the variety of criteria forms a polyhierarchical structure established by the directed non-reflective relation of criteria dependency and called the generating polyhierarchy. Elementary concretization by a single criterion is associated with a logical predicate represented by the respective elementary attribute. Classification categories are implicitly identified as attributive expressions encoding logical compositions of elementary predicates from different criteria. They form the induced polyhierarchy of categories that is established by the directed relation of implication of the respective attributive expressions.

The generating polyhierarchy implicitly and unambiguously defines the induced polyhierarchy, thus making redundant an explicit description of the equivalent DAG. It provides a very compact representation of the target classification, without explicitly enumerating a vast majority of classification categories. In particular, all the intermediate abstract categories emerging when navigating over the polyhierarchy, and initially empty categories to be filled with objects, can be reconstructed dynamically in run time. Operations of the taxonomy algebra such as selection by a superposition of criteria, retrieval of particular sub-trees, test for inclusion, and set-theoretic operations, are executed directly in terms of attributive expressions without ever referring to the classification universe.

## 2. Conjunctive Classifications by Criteria

Let **A** be a finite or infinite set of unspecified objects (universe). We will build a classification of objects **a** ∈ **A** as a hierarchical decomposition of **A** into a system of subsets (*categories* of classification) using a system of concretization rules (*criteria* of classification).

To introduce the notations we will first consider an elementary case, which is a classification by a single criterion. Let's introduce an unambiguous function **attr(a)** on the universe **A** that takes integral values from 1 to N. The function **attr** defines partitioning of **A** into mutually disjoint categories (equivalence classes) **A**(i):

$$\mathbf{A} = \bigcup_{i=1}^{N} \mathbf{A}(i), \quad \mathbf{A}(i) \cap \mathbf{A}(j) = \emptyset \text{ for } i \neq j,$$

where $\mathbf{a} \in \mathbf{A}(i) \Leftrightarrow \mathbf{attr(a)} = i, \quad 1 \leq i \leq N$.

This partitioning is called a *classification by criterion* **C**, criterion **C** being defined by the function **attr**, and the categories **A**(i) are said to be *generated* by the criterion **C**. Distinct values **attr**(**a**) = i are called *branches* of the criterion **C**, and ordered pairs (**C**,i) are called (elementary) *attributes* assigned to elements **a** ∈ **A** by the criterion **C**. The number of branches of a criterion is called its *cardinality*.

**Note 2.1** Numeric identification of criteria branches is used here just for convenience. In practical implementation branches may be represented by any unordered but denumerable collections of distinct symbols, such as keywords, references to database records, etc.

In case of a concurrent application of several classification criteria $C_p$, each criterion is defined by a correspondent unambiguous functions $\text{attr}_p(a)$, $1 \leq p \leq M$. Then we have a system of M independent partitionings of the universe **A** into mutually disjoint categories $A_p(i_p)$. For further considerations it is convenient to represent the functions $\text{attr}_p(a)$ in terms of sets of mutually exclusive predicates $P_p(i_p)$:

$P_p(i_p) \Leftrightarrow \text{attr}_p(a) = i_p \Leftrightarrow a \in A(i)$, $1 \leq i_p \leq N_p$,

$P_p(i_p) \wedge P_p(j_p) = \text{false}$ for $i_p \neq j_p$, (2.1)

where $N_p$ denote cardinalities of the criteria $C_p$.

Building **c**lassifications by a superposition of several criteria is quite obvious. Let's consider inclusion

$a \in A_{\{p(s)\}}\{i_s\} = A_{p(1)}(i_1) \cap A_{p(2)}(i_2) \cap \ldots \cap A_{p(L)}(i_L)$,

$1 \leq L \leq M$, $1 \leq i_s \leq N_{p(s)}$, $1 \leq s \leq L$,

where $\{p(s)\}$ is a subset of L criteria numbers such that $1 \leq p(s) \leq M$ and $p(s) \neq p(t)$ for $s \neq t$, and $\{i_s\}$ is a subset of L corresponding criteria branches. This inclusion means that the element **a** is assigned a set of L respective attributes $\{(C_{p(s)}, i_s)\}$, without regard to all other criteria $C_q$, $q \notin \{p(s)\}$ (if any). Therefore, the superposition of criteria $\{C_{p(s)}, 1 \leq s \leq L\}$ generates partitioning of **A** into $N_{p(1)}N_{p(2)}\ldots N_{p(L)}$ mutually disjoint categories $A_{\{p(s)\}}\{i_s\}$:

$$A = \bigcup_{i_1=1}^{N_{p(1)}} \bigcup_{i_2=1}^{N_{p(2)}} \ldots \bigcup_{i_L=1}^{N_{p(L)}} A_{\{p(s)\}}\{i_s\}, \quad (2.2)$$

$A_{\{p(s)\}}\{i_s\} \cap A_{\{p(s)\}}\{j_s\} = \emptyset$ for $\{i_s\} \neq \{j_s\}$,

where $a \in A_{\{p(s)\}}\{i_s\} \Leftrightarrow f_{\{p(s)\}}\{i_s\} \Leftrightarrow$

$\Leftrightarrow \text{attr}_{p(1)}(a) = i_1, \text{attr}_{p(2)}(a) = i_2, \ldots, \text{attr}_{p(L)}(a) = i_L$,

and

$$f_{\{p(s)\}}\{i_s\} = \bigwedge_{s=1}^{L} P_{p(s)}(i_s). \quad (2.3)$$

Each of partitionings (2.2) uniquely defined by the subset $\{p(s)\}$ represents a partial L-parameter disjunctive classification of the universe **A**. The respective subsets of attributes

$$S_{\{p(s)\}}\{i_s\} = \{(C_{p(s)}, i_s), 1 \leq s \leq L\} \quad (2.4)$$

defining categories $A_{\{p(s)\}}\{i_s\}$ in terms of conjunctive logical functions (2.3) are called *simple collections*.

**Note 2.2** If criteria $C_p$ are semantically related, then some simple collections may correspond to contradictory descriptions of properties of the classified objects. In this case the respective categories are identically empty subsets. But the use of criteria dependency and generalized disjunctive schemes (see sections 3, 5 below) allows to avoid contradictive descriptions.

**Note 2.3** The introduced formalism is also applicable to the cases of infinite denumerable sets of criteria and infinite criteria cardinalities.

## 3. Generating Polyhierarchies of Criteria

In typical applications, the majority of classification criteria are applicable not to the whole universe **A**, but only to some its subsets. The key element of our approach is that criteria domains of definition are explicitly described by attributes from other criteria, i.e. they are themselves categories of classification. The domain of definition of a criterion $C_q$ is called its *root category* and denoted $\text{root}(C_q)$. We will also say that the criterion's root category *introduces* that criterion. Criteria sharing the same root category are not ordered by rank or any other feature.

Let's introduce a directed binary relation of dependence between criteria. We will say that criterion $C_u$ *depends* on criterion $C_v$, $v \neq u$, and use the notation $C_u \subset C_v$, if the simple collection (or a more general attributive expression introduced later) defining category $A_{\{p(s)\}}\{i_s\} = \text{root}(C_u)$ includes an attribute by the criterion $C_v$, i.e. $v \in \{p(s)\}$. Obviously, the dependency relation is non-reflective, that is $C_u \not\subset C_u$, and transitive, that is if $C_u \subset C_v$ and $C_v \subset C_w$ then $C_u \subset C_w$. Combination of these properties guarantees the absence of loops (cyclic paths) in the system of all dependency relations between criteria.

Consider all globally applicable criteria rooting from the whole universe **A**. If one introduces an imaginary criterion $C_0$ generating **A**, then all those criteria become dependent on $C_0$. Therefore, the entire set of criteria becomes a polyhierarchy representable by a connected directed acyclic graph (DAG) with a single root vertex $C_0$.

Since the variety of all simple collections composed of attributes from different criteria exhaustively defines the plurality of all meaningful categories, the polyhierarchy of criteria is called the ***generating polyhierarchy***. However, it should be emphasized that dependencies between criteria impose restrictions on compositions of simple collections. If, for example, $C_u \subset C_v$, and the simple collection defining **root**($C_u$) includes an attribute ($C_v, i_v$), then any attribute from $C_u$ or other criterion depending on $C_u$ can be used only in combination with the ($C_v, i_v$). The elementary predicates (2.1) corresponding to attributes from dependent criteria cannot be used unless the predicate $P_v(i_v)$ is true. Therefore, the generating polyhierarchy is a «skeleton» or template prescribing meaningful combinations of object properties represented in terms of attributes, i.e. logical structure of the target classification.

**Note 3.1** For convenience, it is supposed that classification criteria are partially ordered by their generalities: $C_p \not\subset C_q$ for $p \leq q$. Likewise, the lists of attributes of simple collections (2.4) are supposed to be ordered by criteria numbers, i.e. $p(s) < p(t)$ and $C_{p(s)} \not\subset C_{p(t)}$ for $s < t$.

## 4. Induced Polyhierarchies of Categories

It can be easily observed that the variety of categories implicitly defined by the generating polyhierarchy also forms a polyhierarchical structure called the ***induced polyhierarchy*** of categories. It is established by the directed binary relation of inclusion, and roots from the universe **A**. Inclusion of categories is equivalent to inverse inclusion of the respective simple collections and to implication ($\rightarrow$) of the respective logical functions (2.3):

$$A_{\{p(s)\}}\{i_s\} \subset A_{\{q(t)\}}\{j_t\} \Leftrightarrow \{(C_{p(s)},i_s)\} \supset \{(C_{q(t)},j_t)\} \Leftrightarrow$$
$$\Leftrightarrow ( h_{\{p(s)\}}\{i_s\} \rightarrow h_{\{q(t)\}}\{j_t\} ), \quad 1 \leq s \leq L_1, 1 \leq t \leq L_2.$$

**Note 4.1** In this paper the implication of two logical functions $f_1(a) \rightarrow f_2(a)$, $a \in A$ is understood as the predicate: «$f_2(a)$ = true for all $a$ such that $f_1(a)$ = true».

Let's consider categories related to a given category $A_{\{p(s)\}}\{i_s\}$ by direct or inverse inclusions and differing from it by a single attribute. These categories are called either ***direct parent*** (***base***) or ***direct child*** (***derived***) categories of $A_{\{p(s)\}}\{i_s\}$, depending on the direction of inclusion. To illustrate the ways of operating with attributive representations of categories, three particular tasks are discussed below: 1) finding all direct parents of a given category; 2) finding all direct children of a given category; and 3) testing two given categories for inclusion.

**4.1. Finding direct parents.** Let's consider any given category $A_{\{p(s)\}}\{i_s\}$ defined by nonempty simple collection $\{(C_{p(s)}, i_s), 1 \leq s \leq L\}$. Obviously, the subset of criteria $\{C_{p(s)}, 1 \leq s \leq L\}$ form a sub-hierarchy with the same imaginary root criterion $C_0$ as the whole generating polyhierarchy. Therefore, that sub-hierarchy must contain at least one criterion $C_{p(m)}$, $1 \leq m \leq L$, called *leaf criterion* of the category $A_{\{p(s)\}}\{i_s\}$, such that $C_{p(s)} \not\subset C_{p(m)}$ for $s = 1,2,\ldots,L$.

Since excluding leaf criterion $C_{p(m)}$ from the considered sub-hierarchy does not impair dependencies between all remaining criteria, it results in a reduced sub-hierarchy of L-1 criteria. Hence, the reduced simple collection $\{(C_{q(t)}, k_t), 1 \leq t \leq L-1\} = \{(C_{p(s)}, i_s), 1 \leq s \leq L, s \neq m\}$ with one less attribute is valid. It defines a direct parent category $A_{\{q(t)\}}\{k_t\}$ including the given one, i.e. $A_{\{p(s)\}}\{i_s\} \subset A_{\{q(t)\}}\{k_t\} \Leftrightarrow \{(C_{p(s)}, i_s)\} \supset \{(C_{q(t)}, k_t)\}$. Thus, for any category $A_{\{p(s)\}}\{i_s\} \neq A$, there exists a set of immediate parents, their number being equal to the number of the category's leaf criteria.

**4.2. Finding direct children.** Let's call *free criteria* of a given category $A_{\{p(s)\}}\{i_s\}$ those criteria $C_f$ that are defined for that category but not used in any of attributes containing in its simple collection, i.e. $A_{\{p(s)\}}\{i_s\} \subset$ **root**($C_f$) while $f \notin \{p(s)\}$. Evidently, the sets of leaf and free criteria of a given category do not overlap. By adding one of attributes ($C_f, i_f$) ($1 \leq i_f \leq N_f$) from the free criterion $C_f$ to the simple collection of $A_{\{p(s)\}}\{i_s\}$ we get an extended simple collection $\{(C_{r(t)}, n_t), 1 \leq t \leq L+1\} = \{\{(C_{p(s)}, i_s), 1 \leq s \leq L\}, (C_f, i_f)\}$ with one more attribute. It defines a direct child category $A_{\{r(t)\}}\{n_t\}$ included in the given one i.e. $A_{\{p(s)\}}\{i_s\} \supset A_{\{r(t)\}}\{n_t\} \Leftrightarrow \{(C_{p(s)}, i_s)\} \subset \{(C_{r(t)}, n_t)\}$. Thus, for any given category $A_{\{p(s)\}}\{i_s\}$ with a non-empty set of free criteria there exists a set of immediate children and their number equals the sum of cardinalities of free criteria of the given category.

**4.3. Test for inclusion.** The problem of testing two given categories $A_{\{p(s)\}}\{i_s\}$ ($1 \leq s \leq L_1$) and $A_{\{q(t)\}}\{j_t\}$ ($1 \leq t \leq L_2$) for inclusion is equivalent to a trivial check of the respective simple collections for the inverse inclusion: $A_{\{p(s)\}}\{i_s\} \subset A_{\{q(t)\}}\{j_t\} \Leftrightarrow \{(C_{p(s)}, i_s)\} \supset \{(C_{q(t)}, j_t)\}$, i.e. $L_1 \geq L_2$, and $p(s) = q(s)$, $i_s = j_s$ for $s = 1,2,\ldots,L_2$.

Generating polyhierarchy together with the lists of criteria branches implicitly describes the target polyhierarchical classification, thus making redundant an explicit enumeration of a vast majority of categories. For practical applications it is a critical question how many categories have to be explicitly enumerated and permanently stored in the form of simple collections or more general attributive expressions (see below). Apparently, for an effective work with the induced polyhierarchy the storage of only 1) root categories defining the structure of generating polyhierarchy, and 2) non-empty categories used as containers for classified objects, is sufficient. All other categories required for navigating over the polyhierarchy, retrieving particular sub-trees, etc., can be dynamically restored in run-time using the generating polyhierarchy.

## 5. Generalized Classification Scheme

The introduced formalism gives a general tool for building, maintaining, and using purely conjunctive polyhierarchical classifications. Because of using representations in terms of simplest logical functions (2.3), it allows to generate categories by only decompositions by criteria and intersections of more general categories. Categories of the induced polyhierarchy form a semi-ring in the set-theoretic sense. However, some practical problems require support for more advanced operations, such as unification and complement. This section describes a generalized version of the method, supporting a full set of set-theoretic operations, while keeping persistent polyhierarchical structures of both variety of criteria, and induced system of categories.

As it is known, a semi-ring of subsets can be complemented to a ring by adding the operation of unification of subsets. Therefore, introduction of a formalism allowing definition of categories as unions of any other categories is sufficient for enabling set-theoretic operations. This can be done by replacing the conjunctive functions (2.3) with more general compositions of elementary predicates (2.1) in the form:

$$\mathbf{d}_{\{p(s,k)\}}\{i_{s,k}\} = \bigvee_{k=1}^{K} \mathbf{h}_{\{p(s,k)\}}\{i_{s,k}\} \quad (5.1)$$

where $\mathbf{h}_{\{p(s,k)\}}\{i_{s,k}\} = \bigwedge_{s=1}^{L_k} P_{p(s,k)}(i_{s,k})$, $1 \leq k \leq K$.

The terms $\mathbf{h}_{\{p(s,k)\}}\{i_{s,k}\}$ in logical polynomials (5.1) are conjunctive logical functions corresponding to simple collections, likewise functions (2.3). Each polynomial defines category $\mathbf{A}_{\{p(s,k)\}}\{i_{s,k}\} = \mathbf{A}(\mathbf{d}_{\{p(s,k)\}}\{i_{s,k}\})$ as a subset of all elements $\mathbf{a} \in \mathbf{A}$ such that

$$\mathbf{a} \in \mathbf{A}_{\{p(s,k)\}}\{i_{s,k}\} \Leftrightarrow \mathbf{d}_{\{p(s,k)\}}\{i_{s,k}\}.$$

For any two polynomials $\mathbf{d}_1$ and $\mathbf{d}_2$

$$\mathbf{A}(\mathbf{d}_1) \subset \mathbf{A}(\mathbf{d}_2) \Leftrightarrow (\mathbf{d}_1 \to \mathbf{d}_2), \quad (5.2)$$

$$\mathbf{A}(\mathbf{d}_1 \vee \mathbf{d}_2) = \mathbf{A}(\mathbf{d}_1) \cup \mathbf{A}(\mathbf{d}_2), \quad (5.3)$$

$$\mathbf{A}(\mathbf{d}_1 \wedge \mathbf{d}_2) = \mathbf{A}(\mathbf{d}_1) \cap \mathbf{A}(\mathbf{d}_2), \quad (5.4)$$

$$\mathbf{A}(\mathbf{d}_1 \wedge -\mathbf{d}_2) = \mathbf{A}(\mathbf{d}_1) \setminus \mathbf{A}(\mathbf{d}_2), \quad (5.5)$$

where «-» denotes logical negation. We will call the categories defined by logical polynomials (5.1) *composite* ones to distinguish them from *simple* categories defined by the conjunctive functions (2.3).

Representing categories in terms of polynomials (5.1) imposes no restriction on the use of composite categories as roots for criteria, thus preserving the meaning of the dependency relations between them. Therefore, composite categories can be used in the construction of the generating polyhierarchy. However, since it is built without referring to actual objects composing the universe **A**, the meaning of the relations (5.2) - (5.5) in the context of our method should be clarified.

First, classification categories are treated as imaginary subsets of all potentially existing objects with combinations of properties permitted by the construction of the generating polyhierarchy. Second, any relationships between categories stipulated only by «external» semantics, but not reflected in the structure of the generating polyhierarchy, are excluded from consideration. Third, all set-theoretic operations on categories and relations between them are required to be invariant with respect to increasing criteria cardinalities and introducing new criteria.

In practical implementations of the generalized scheme, logical polynomials (5.1) can be encoded by the attributive assemblies

$$\mathbf{d}_{\{p(s,k)\}}\{i_{s,k}\} \sim \{\mathbf{S}_{\{p(s,k)\}}\{i_{s,k}\}, 1 \leq k \leq K\}, \quad (5.6)$$

where «~» denote correspondence between logical function and its attributive representation. The subsets of attributes

$$\mathbf{S}_{\{p(s,k)\}}\{i_{s,k}\} = \{(\mathbf{C}_{p(s,k)}, i_{s,k}), 1 \leq s \leq L_k\} \sim \mathbf{h}_{\{p(s,k)\}}\{i_{s,k}\},$$
$$p(s,k) \neq p(t,k) \text{ for } s \neq t, \quad 1 \leq k \leq K, \quad (5.7)$$

are simple collections encoding conjunctive terms of logical polynomials likewise (2.4). Without loss of generality we can assume that none of the simple collections (5.7) includes another, i.e. $\mathbf{S}_{\{p(s,k)\}}\{i_{s,k}\} \not\subset \mathbf{S}_{\{p(s,\ell)\}}\{i_{s,\ell}\}$ for $k \neq \ell$. The attributive representations (5.6) called *unions of simple collections*, by their definition imply the conjunctive composition of object properties defined by attributes within each simple collection (5.7), and the disjunctive composition of the sets of properties defined by separate simple collections.

Computing the complements of categories considered in section 6 below requires an expression for the negation of a logical polynomial:

$$-\mathbf{d}_{\{p(s,k)\}}\{i_{s,k}\} = \bigwedge_{k=1}^{K} (-\mathbf{h}_{\{p(s,k)\}}\{i_{s,k}\}) = \quad (5.8)$$

$$= \bigwedge_{k=1}^{K} \bigvee_{s=1}^{L_k} ( ( \bigwedge_{t=1}^{s-1} \mathbf{P}_{p(t,k)}(i_{t,k}) ) \wedge (-\mathbf{P}_{p(s,k)}(i_{s,k})) ),$$

where predicates $\mathbf{P}_{p(s,k)}(i_{s,k})$ are supposed to be partially ordered by generality of the respective criteria (see note 3.1). The additional cofactors ($\wedge \mathbf{P}_{p(t,k)}(i_{t,k})$, $1 \leq t \leq s-1$) are included in (5.8) for a proper representation of domains of applicability of predicates $\mathbf{P}_{p(s,k)}(i_{s,k})$. In practical implementations, negations of predicates can be encoded by complements of the respective attributes

$$\mathbf{compl}(\mathbf{C}_{p(s,k)}, i_{s,k}) \sim -\mathbf{P}_{p(s,k)}(i_{s,k}) = \bigvee_{j \neq i_{s,k}} \mathbf{P}_{p(s,k)}(j) \quad (5.9)$$

defined on $\mathbf{root}(\mathbf{C}_{p(s,k)})$. Whenever necessary, the complements can be excluded from attribute collections by using distributivity of the operations $\wedge$ and $\vee$.

## 6. The Taxonomy Algebra

The generalized formalism allows executing set-theoretic operations on categories directly in terms of unions of simple collections. This section briefly discusses basic algorithms of the *taxonomy algebra* that may be required when working with the induced polyhierarchy. They include: 1) test for inclusion, 2) computing union, 3) computing intersection, 4) computing complement (difference), 5) retrieving direct parent and direct child categories.

In this section, details of operations on simple collections already discussed in section 4 above are omitted. Simple categories defined by simple collections (5.7) of unions (5.6) are called ***union components***.

**6.1. Test for inclusion.** The relation of inclusion between categories is equivalent to the implication of their logical polynomials (see (5.2) and note 4.1). Due to the logical independence of predicates (2.1) from different criteria, none of the polynomials (5.1) for $K \geq 2$ can be represented as a conjunction of predicates. Hence, for a set of simple categories $\mathbf{A}_k$ ($1 \leq k \leq K$), such that $\mathbf{A}_k \not\subset \mathbf{A}_\ell$ for $k \neq \ell$, and a simple category $\mathbf{B} \subset \mathbf{A}_1 \cup \mathbf{A}_2 \cup ... \cup \mathbf{A}_K$, there exists a number $m$ ($1 \leq m \leq K$) such that $\mathbf{B} \subset \mathbf{A}_m$.

Let's consider two arbitrary composite categories represented by unions of simple collections:

$$\mathbf{A}_{\{p(s,k)\}}\{i_{s,k}\} \sim \{\mathbf{S}_{\{p(s,k)\}}\{i_{s,k}\}, 1 \leq k \leq K_1\} \text{ and}$$

$$\mathbf{A}_{\{q(t,m)\}}\{i_{t,m}\} \sim \{\mathbf{S}_{\{q(t,m)\}}\{j_{t,m}\}, 1 \leq m \leq K_2\}. \quad (6.1.1)$$

For $\mathbf{A}_{\{p(s,k)\}}\{i_{s,k}\} \subset \mathbf{A}_{\{q(t,m)\}}\{i_{t,m}\}$, it is necessary and sufficient that each of the components of the first union were included into some component of the second union:

$$\forall k \, (1 \leq k \leq K_1) \, \exists m = m(k) \, (1 \leq m \leq K_2):$$

$$\mathbf{S}_{\{p(s,k)\}}\{i_{s,k}\} \supset \mathbf{S}_{\{q(t,m)\}}\{j_{t,m}\}. \quad (6.1.2)$$

**6.2. Computing union.** The algorithm is based on the formula (5.3). The union of two given composite categories (6.1.1) is computed by merging the lists of union components:

$$\mathbf{A}_{\{p(s,k)\}}\{i_{s,k}\} \cup \mathbf{A}_{\{q(t,m)\}}\{i_{t,m}\} \sim \quad (6.2.1)$$

$$\sim \{\{\mathbf{S}_{\{p(s,k)\}}\{i_{s,k}\}, 1 \leq k \leq K_1\}, \{\mathbf{S}_{\{q(t,m)\}}\{j_{t,l}\}, 1 \leq m \leq K_2\}\}$$

with the subsequent removal of redundancy, i.e. ***reduction*** of the description. Reduction means removing all the union components $\mathbf{S}$ such that the resulting union of simple collections (6.2.1) includes another component $\mathbf{T} \supset \mathbf{S}$.

**6.3. Computing intersection.** This algorithm is based on the formula (5.4). The intersection of two given composite categories (6.1.1) equals the union of all non-empty pair-wise intersections of the union components:

$$\mathbf{A}_{\{p(s,k)\}}\{i_{s,k}\} \cap \mathbf{A}_{\{q(t,m)\}}\{i_{t,m}\} \sim \quad (6.3.1)$$

$$\sim \{\mathbf{T}_{k,m}, 1 \leq k \leq K_1, 1 \leq m \leq K_2, \mathbf{T}_{k,m} \neq \mathsf{null}\},$$

where

$\mathbf{T}_{k,m} = \mathbf{S}_{\{p(s,k)\}}\{i_{s,k}\}$ if $\mathbf{S}_{\{q(t,m)\}}\{j_{t,m}\} \subset \mathbf{S}_{\{p(s,k)\}}\{i_{s,k}\}$,

$\mathbf{T}_{k,m} = \mathbf{S}_{\{q(t,m)\}}\{j_{t,m}\}$ if $\mathbf{S}_{\{p(s,k)\}}\{i_{s,k}\} \subset \mathbf{S}_{\{q(t,m)\}}\{j_{t,m}\}$,

$\mathbf{T}_{k,m} = \mathsf{null}$ if $\mathbf{S}_{\{q(t,m)\}}\{j_{t,m}\} \not\subset \mathbf{S}_{\{p(s,k)\}}\{i_{s,k}\}$
and $\mathbf{S}_{\{p(s,k)\}}\{i_{s,k}\} \not\subset \mathbf{S}_{\{q(t,m)\}}\{j_{t,m}\}$.

Here «null» denotes a simple collection containing more than one attribute by the same criterion and corresponding to an identically empty category. The resulting union of simple collections $\{\mathbf{T}_{k,m} \neq \mathsf{null}\}$ should be reduced (see subsection 6.2).

**6.4. Computing complement.** This algorithm is based on (5.5), (5.8) and (5.9). Simple transformations of the attributive representations (6.1.1) result in the expression

$$\mathbf{A}_{\{p(s,k)\}}\{i_{s,k}\} \setminus \mathbf{A}_{\{q(t,m)\}}\{i_{t,m}\} =$$

$$= \mathbf{A}_{\{p(s,k)\}}\{i_{s,k}\} \cap ( \bigcap_{m=1}^{K_2} \bigcup_{t=1}^{L_{2,m}} \mathbf{B}_{t,m} ), \quad (6.4.1)$$

where $L_{2,m}$ ($1 \leq m \leq K_2$) are total numbers of attributes in simple collections $\mathbf{S}_{\{q(t,m)\}}\{j_{t,m}\}$, and the ancillary categories $\mathbf{B}_{m,t}$ are defined by unions of $N_{q(t,m)}-1$ simple collections $\mathbf{T}_{t,m,r}$:

$$B_{t,m} \sim \{T_{t,m,r}, 1 \leq r \leq N_{q(t,m)}, r \neq j_{t,m}\}, \quad (6.4.2)$$

$$T_{t,m,r} = \{\{(C_{q(n,m)}, j_{n,m}), 1 \leq n \leq t-1\}, (C_{q(t,m)}, r)\},$$

$$1 \leq t \leq L_{2,m\ 2}, \ 1 \leq m \leq K_2,$$

where $N_{q(t,m)}$ are cardinalities of the criteria $C_{q(t,m)}$. Combination of expressions (6.4.1) and (6.4.2) allows to compute complement as a superposition of unifications and intersections of categories $B_{t,m}$ using the algorithms (6.2.1) and (6.3.1). In a general case, direct use of these formulae may prove to be costly, but possible ways of optimization are quite obvious.

Since the set of simple collections $T_{t,m,r}$ depends on the cardinalities $N_{q(t,m)}$, the complement operation in the given formulation is not invariant with respect to increasing criteria cardinalities. However it becomes invariant if the notion of simple collection is generalized by allowing it to include complements of elementary attributes. In that case simple collections would denote conjunctive compositions of both elementary predicates and their negations (see 5.9), thus allowing to define $B_{t,m}$ from (6.4.1) as:

$$B_{t,m} \sim \{(C_{q(n,m)}, j_{n,m}), 1 \leq n \leq t-1\}, \mathbf{compl}(C_{q(t,m)}, j_{t,m})\},$$

$$1 \leq t \leq L_{2,m}, \ 1 \leq m \leq K_2.$$

This generalization translates into some trivial modifications of the algorithms of subsections 6.1 – 6.3.

**6.5. Retrieving direct parents and children.** It is natural to call *direct parent* (*base*) and *direct child* (*derived*) categories of a given category **E** those categories $B \supset E$ and $D \subset E$, that result from **E** after performing a single elementary generalization and, respectively, specialization. By «elementary» generalizations and specializations we mean those that cannot be represented as compositions of simpler operations. More exactly, there must be no intermediate categories **B\*** and **D\*** such that **B\*** ≠ **E**, **B\*** ≠ **B**, $B \supset B^* \supset E$ and **D\*** ≠ **E**, **D\*** ≠ **D**, $D \subset D^* \subset E$, respectively.

In semantics of the formalism of unions of simple collections elementary generalizations and specializations correspond to additions and subtractions of various simple categories without free criteria, i.e. leaf categories. Therefore, direct parents and children of a given category **E** are all non-empty categories **E** \ **F** and **E** ∪ **G**, respectively, where $F \subset E$ and $G \not\subset E$ are leaf categories of the induced polyhierarchy.

## 7. Attributive Expressions

It can be seen that our approach permits different ways of implementation depending on semantics of the subsets of attributes identifying categories. The semantics of simple collections discussed in sections 2 – 4 above describes only a purely conjunctive classifications with a limited set of set-theoretic operations. In contrast, the semantics of unions of simple collections described in sections 5 and 6 can be used for constructing very complex polyhierarchies with freely combined conjunctive and disjunctive operations, at the expence of performance. Therefore, the way of encoding logical functions defining categories should be chosen depending on the required functionality of target classification.

The customizable information structures defining categories in terms of logical compositions of the elementary predicates (2.1) are called *attributive expressions*. Besides simple collections and unions of simple collections, which are particular cases of attributive expressions, some other ways of composing predicates can be also relevant in certain practical cases.

For example, the full support of set-theoretic operations provided by the semantics of unions of simple collections may be redundant for many applications. For the majority of practical needs it is sufficient to provide only the option of disjunction of predicates corresponding to branches of the same criterion. For these cases, the following logical functions provide an appropriate representation of categories:

$$c_{\{p(s)\}}\{i_{s,k}\} = \bigwedge_{s=1}^{M} u_{p(s)}\{i_{s,k}\}, \quad (7.1)$$

where $\quad u_{p(s)}\{i_{s,k}\} = \bigvee_{k=1}^{K_s} P_{p(s)}(i_{s,k}), \quad 1 \leq s \leq M.$

Those representations can be encoded by the attributive assemblies

$$c_{\{p(s,k)\}}\{i_{s,k}\} \sim \{U_{p(s)}\{i_{s,k}\}, 1 \leq s \leq M\}, \quad (7.2)$$

where the subsets of attributes

$$U_{p(s)}\{i_{s,k}\} = \{(C_{p(s)}, i_{s,k}), 1 \leq k \leq K_s\} \sim u_{p(s)}\{i_{s,k}\}, \quad (7.3)$$

$i_{s,k} \neq i_{s,\ell}$ for $k \neq \ell.$,

encoding disjunctive terms of the functions (7.1) are called *branch unions* of the respective criteria $C_{p(s)}$. The attributive representations (7.2) are called *collections with branch unions*. By definition they imply the disjunctive composition of object properties defined by attributes within each branch union (7.3), and the conjunctive composition of the sets of properties defined by separate branch unions. Simple collections (2.4) are equivalent to collections with brunch unions (7.2) where each union has cardinality 1.

Using mutual distributivity of operations ∧ and ∨ any of the functions (7.1) can be transformed to the polynomial form (5.1). However, the opposite conversion, requiring a complete factorization of the polynomial (5.1), is not generally possible. Therefore, the functions (7.1) form an intermediate class of compositions of the predicates (2.1) between conjunctive representations (2.3) and polynomials (5.1).

The semantics of collections with branch unions allows definition of categories using decompositions by criteria, intersections of more general categories, and unifications of categories generated by branches of the same criterion. The algorithms for testing given categories for inclusion and computing intersections in terms of attributive expressions (7.2) are obvious enough to omit their discussion. However, the procedures of retrieving direct parent (base) and child (derived) categories of a given category require some additional consideration.

**7.1. Finding direct children.** Let's consider a category $\mathbf{A}_{\{p(s)\}}\{i_{s,k}\}$ defined by a collection (7.2). If some branch unions $\mathbf{U}_{p(s)}\{i_{s,k}\}$ of that collection have cardinality $K_s \geq 2$, then removing from such a union one of its attributes $(\mathbf{C}_{p(s)}, i_{s,n})$ ($1 \leq n \leq K_s$) results in a non-empty reduced branch union of cardinality $K_s-1$. Since the resulting reduced collection remains within domains of definition of criteria $\mathbf{C}_{p(s)}$ ($1 \leq s \leq M$), it defines a valid category included into $\mathbf{A}_{\{p(s)\}}\{i_{s,k}\}$ and differing from it by a single attribute $(\mathbf{C}_{p(s)}, i_{s,n})$, i.e. a direct child category. The total number of non-empty children equals $\Sigma K_s - M$.

The way of specialization by adding a free attribute to a simple collection discussed in subsection 4.2 can be represented as a superposition of eliminations of attributes from branch unions. In fact, any free criterion $\mathbf{C}_f$ of a simple collection can be explicitly represented by its **total branch union** $\mathbf{U}_f = \{(\mathbf{C}_f, i), 1 \leq i \leq N_f\}$, meaning absence of any specialization by that criterion. Hence, any simple collection is equivalent to a collection with branch unions composed of total unions from free criteria and elementary attributes from other ones. A sequence of $N_f-1$ eliminations of attributes from a total branch union $\mathbf{U}_f$ is equivalent to adding a single attribute from the $\mathbf{C}_f$.

**7.2. Finding direct parents.** The direct parent categories of a given category $\mathbf{A}_{\{p(s)\}}\{i_{s,k}\}$ defined by a collection (7.2), are retrieved by adding attributes to the brunch unions (7.3). However, one should first reveal the attributes that can be added to the collection without violating domains of definition of the participating criteria. Let's define the **hull** of a category $\mathbf{A}_{\{p(s)\}}\{i_{s,k}\}$ as the most general category providing applicability of all the criteria $\mathbf{C}_{p(s)}$ ($1 \leq s \leq M$):

$$\mathbf{hull}(\mathbf{A}_{\{p(s)\}}\{i_{s,k}\}) = \bigcap_{s=1}^{M} \mathbf{root}(\mathbf{C}_{p(s)}). \qquad (7.2.1)$$

The attributes that can be added to brunch unions (7.3) of the collection (7.2) without exceeding bounds of the hull (7.2.1) are called **hull-compatible attributes** of the category $\mathbf{A}_{\{p(s)\}}\{i_{s,k}\}$. Extension of any of the brunch unions (7.3) by a single hull-compatible attribute results in a collection with branch unions defining direct parent category.

Evidently, the way of generalization by removing a leaf criterion's attribute from a simple collection discussed in subsection 4.1, can be represented as a superposition of additions of hull-compatible attributes to branch unions. Since all the attributes of a leaf criterion are hull-compatible, they can be successively added to the respective branch union until it becomes the total one. The latter means no specialization by the respective criterion; hence it can be excluded from the collection.

If the target classification extensively uses unions of both criteria branches and arbitrary categories, then combined attributive expressions can be useful. In particular, to reduce accumulation of conjunctive terms in the polynomials (5.1) one can define categories by logical functions of the following form:

$$\mathbf{e}_{\{p(s,r)\}}\{i_{s,k,r}\} = \bigvee_{r=1}^{R} \mathbf{g}_{\{p(s,r)\}}\{i_{s,k,r}\}, \qquad (7.4)$$

where $\mathbf{g}_{\{p(s,r)\}}\{i_{s,k,r}\} = \bigwedge_{s=1}^{M_r} \bigvee_{k=1}^{K_{s,r}} \mathbf{P}_{p(s,r)}(i_{s,k,r}).$

Since the terms $\mathbf{g}_{\{p(s,r)\}}\{i_{s,k,r}\}$ are similar to the functions (7.1) by their structure, the respective attributive expressions may be called **unions of collections with branch unions**. Because of mutual distributivity of operations $\wedge$ and $\vee$ the functions (7.4) do not have unique representations, thereby requiring selection of their appropriate application-specific canonical form.

## 8. Practical Implementation

Building a polyhierarchical classification starts from the selection of a system of criteria allowing to distinguish classified objects by their meaningful properties. At this stage one has to determine which form of the attributive expressions is sufficient to support the required functionality, while keeping in mind that the domains of criteria applicability must be themselves categories of the target classification.

The formalism of simple collections described in sections 2 – 4 is adequate for purely conjunctive schemes that do not require definition of categories via unification of object properties. In more complex cases either collections with branch unions (see section 7) or unions of simple collections (see sections 5 and 6) can be used. The classification developer can design yet another form of attributive expressions encoding application-specific logical compositions of the predicates (2.1). Since the amount of required computer resources depends on generality of attributive expressions, their optimal form should be carefully weighted.

At the next stage the generating polyhierarchy should be constructed by exactly defining domains of criteria applicability. It may be described in terms of either the attributive expressions explicitly representing root categories, or any other equivalent form allowing an unambiguous retrieval of root categories. In particular, for many applications it is useful to represent the generating hierarchy in a most observable form for its better comprehension. Only criteria branches used in definitions of the domains of criteria applicability need to be enumerated at this stage, as all other branches can be added when further categorizing particular objects. After the generating polyhierarchy is constructed, the induced polyhierarchy is implicitly and completely defined.

In should be emphasized that the generating polyhierarchy is a self-consistent, compact, portable, and re-usable information structure serving as a template classification. It can be further associated with one or more particular sets of objects, included in more general classifications as a standard component, or used as a prototype for more comprehensive classifications. Therefore, its construction can be completely separated from the final stage discussed below, while the latter may be repeated with different sets of classified objects.

At the final stage the generating polyhierarchy is incorporated into a client database and associated with available objects. The structure of relations between auxiliary tables used to store the generating polyhierarchy should reflect semantics of the selected form of attributive expressions. Typically, a full support of classification functionality can be achieved by adding to the database only five or six interrelated tables, such as «Criteria», «Branches», «Attributes», «Simple Collections», «Collection Unions», and «Categories». The latter can be directly referenced from the client repositories.

When associating classified objects with the implicitly defined induced polyhierarchy, all the additional descriptive data, such as persistent attributive expressions defining non-empty categories and references to them, should be recorded to the database. This is a quite formal, lockstep procedure.

When practically implementing our approach it is worth to consider the optimization and extension techniques mentioned below:

- storage of persistent attributive expressions in a specifically designed compressed form that allows to directly perform massive operations without expanding the expressions;

- definition of categories in terms of relative attributive expressions to be implicitly merged with predefined persistent reference-point expressions;

- introduction of criteria defining domains of applicability of supplementary search tools, such as parameter range and keyword search engines;

- permanent storage of the most abstract identically empty categories, arising from the use of semantically related criteria (if any), to facilitate detection of contradictory queries;

- development of an interface protocol allowing to hide an actual structure of the generating polyhierarchy to make it possible to use some conventional or standard systems of notions, terms, queries, etc.

The described formalism is used at QNT Software Development for the classification of mathematical objects and their computer representations. The primary requirement to that classification was an option to compare the objects by generality to make it possible to check consistency of software components in a cross-language environment. A relatively small fragment of the respective generating polyhierarchy is shown in Figure 1 as an example.

## 9. Advantages of the Method

The presented formalism provides a number of advantages over commonly used tree-structured and faceted schemes. The most important of them are briefly discussed below

**9.1. Compactness and uniformity**. Descriptive data structures defining a classification by our method have by an order of magnitude smaller size compared with equivalent descriptions in terms of trees, forests, and general DAGs. In particular, neither intermediate abstract nor empty leaf categories have to be enumerated or stored. Therefore, the total number of permanently stored attributive expressions representing persistent categories never exceeds the number of used criteria plus the number of classified objects (see section 4). The structure of descriptive data is simple and uniform, because it does not use any auxiliary multi-level constructions like aggregations, compositions, sub- and meta-facets, their purposes, roles, etc. Thus, our approach can be easily implemented in a standard DBMS environment with no or minimal additional programming. At the same time it does not restrict functionality of a classification, since 1) many relations between categories can be expressed in terms of dependent criteria, and 2) the formalism does not prevent introduction of relations with other polyhierarchies (if any), or combination with other classification techniques.

**9.2. Unambiguousness and consistency.** Since the generating polyhierarchy uniquely defines the structure of classification, it eliminates a lot of secondary issues requiring heuristic considerations. For example, the developer does not have to keep in mind how to distribute different parts of the classification over abstraction levels (meta-facets), how many categories and relations between them have to be permanently stored, how to rank search options by their significance, and the like. In addition, a properly designed generating polyhierarchy provides an automatic control of consistency of the attributes, thus preventing erroneous categorization of particular objects.

**9.3. Flexibility.** The implicit description of a classification by its generating polyhierarchy dramatically simplifies unplanned modifications when developing and further refining a classification. For example, extension of the induced polyhierarchy, required for taking into account new selection options, can be done trivially by adding new branches to criteria. Generalizations and refinements are performed by introducing new criteria in the existent generating polyhierarchy. More complex operations, such as composing several classifications into one, can be done automatically by merging the respective attributive expressions.

**9.4. Computational efficiency.** Attributive expressions define absolute positions of corresponding categories in the global polyhierarchy rather than local relations between them. This drastically simplifies performing essentially non-local tasks, such as test for inclusion, and determining the nearest common parents and children of a given set of categories. Our method reduces the respective algorithms to simple logical operations on attributive expressions (see sections 6 and 7), thus requiring neither combinatorial search nor storage of additional descriptions.

## 10. Conclusion

A new general formalism for describing and manipulating complex polyhierarchical information structures is developed. It is based on a concise system of primary notions providing a straightforward and mathematically rigorous approach to the construction of real-world multi-criteria classifications. The formalism satisfies the six basic requirements listed in section 1, thus exceeding other classification schemes by its functionality, flexibility, and range of applicability. It allows an efficient implementation in common DBMS, while considerably reducing the amount of computer resources required for storage, maintenance, and use of complex hierarchical classifications. The developed formalism can be implemented in any applications that use hierarchically structured information, such as

- taxonomical and content management systems;
- expert, artificial intelligence, and machine learning systems;
- intelligent control systems and robots;
- data and knowledge bases;
- internet search engines, online documentation, and help subsystems;
- application-specific lists, catalogues, and directories;
- compilers for object- and aspect-oriented languages with multiple inheritance;
- generative and intentional programming environments;
- components of operating systems (file and folder catalogues, registry, etc.);
- component based software engineering systems.

## 11. Acknowledgement

This work has been supported by the Coalco AG (Switzerland).

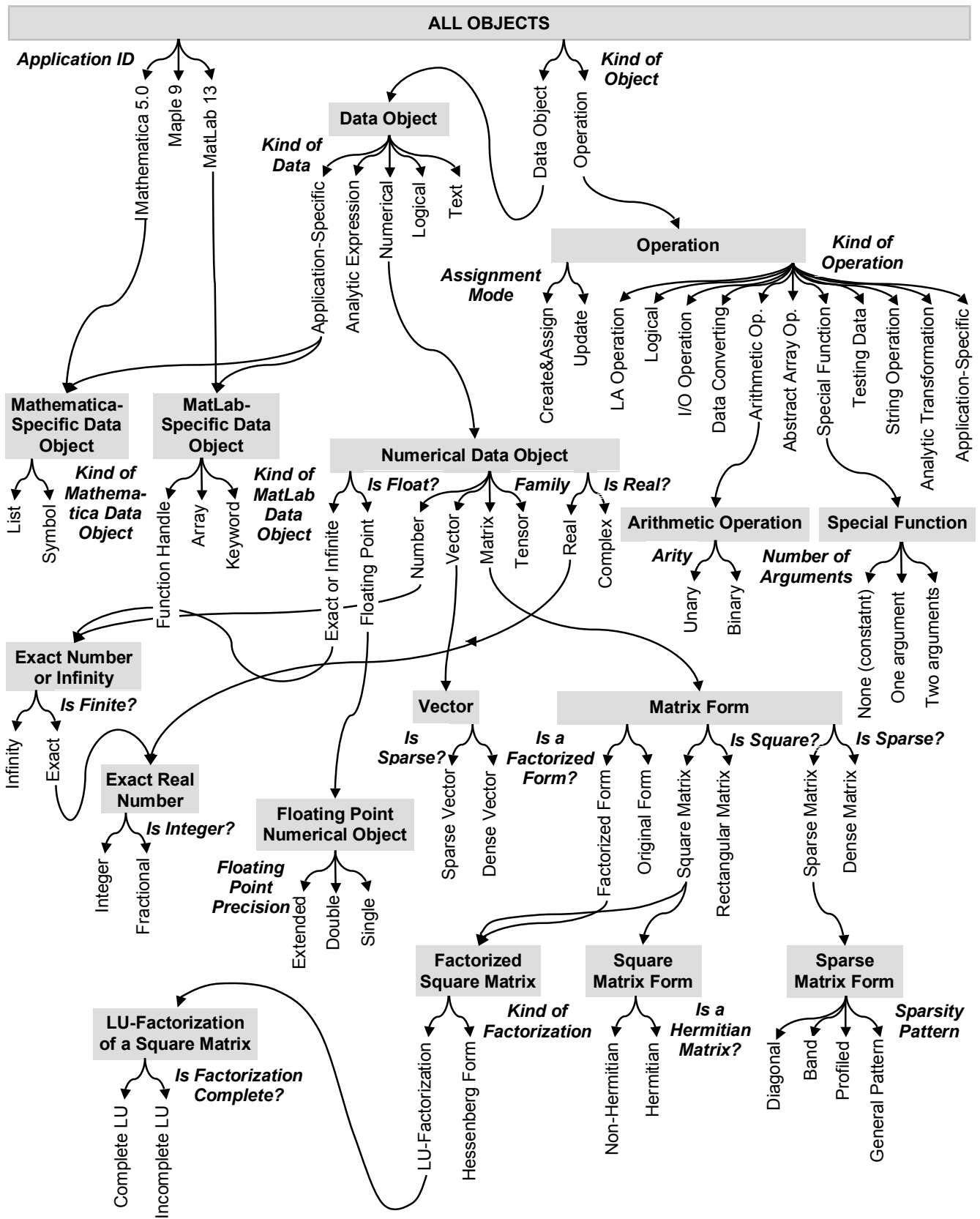

Figure 1. A fragment of the generating polyhierarchy of a classification of mathematical objects. Root categories of criteria are shown as grey blocks.